\documentclass[10pt, a4paper]{article}
\usepackage{lrec2022} 
\usepackage{multibib}
\newcites{languageresource}{Language Resources}
\usepackage{graphicx}
\usepackage{tabularx}
\usepackage{soul}
\usepackage{titlesec}
\titleformat{\section}{\normalfont\large\bfseries\center}{\thesection.}{1em}{}
\titleformat{\subsection}{\normalfont\SmallTitleFont\bfseries\raggedright}{\thesubsection.}{1em}{}
\titleformat{\subsubsection}{\normalfont\normalsize\bfseries\raggedright}{\thesubsubsection.}{1em}{}
\renewcommand\thesection{\arabic{section}}
\renewcommand\thesubsection{\thesection.\arabic{subsection}}
\renewcommand\thesubsubsection{\thesubsection.\arabic{subsubsection}}

\usepackage{epstopdf}
\usepackage[utf8]{inputenc}

\usepackage{hyperref}
\usepackage{xstring}
\usepackage{booktabs}
\usepackage{multirow}
\usepackage{caption}
\usepackage{arabtex}
\usepackage{float}
\usepackage{utf8}
\setcode{utf8}

\usepackage{color}
\usepackage{pifont}

\newcommand{\cmark}{\ding{51}}%
\newcommand{\xmark}{\ding{55}}%

\definecolor{light-gray}{gray}{0.95}
\newcommand{\code}[1]{\colorbox{light-gray}{\texttt{#1}}}

\title{Meta AI at Arabic Hate Speech 2022: \\ MultiTask Learning with Self-Correction for Hate Speech Classification}

\name{Badr AlKhamissi, Mona Diab} 

\address{Responsible AI, Meta \\
         Seattle, USA \\
        \{bkhmsi, mdiab\}@fb.com\\}

\abstract{
    In this paper, we tackle the Arabic Fine-Grained Hate Speech Detection shared task and demonstrate significant improvements over reported baselines for its three subtasks. The tasks are to predict if a tweet contains (1) \textit{Offensive} language; and whether it is considered (2) \textit{Hate Speech} or not and if so, then predict the (3) \textit{Fine-Grained Hate Speech} label from one of six categories. Our final solution is an ensemble of models that employs multitask learning and a self-consistency correction method yielding 82.7\% on the hate speech subtask---reflecting a 3.4\% relative improvement compared to previous work. 
}

\begin{document}

\maketitleabstract

\section{Introduction}

\textbf{Disclaimer}: {\color{red} \textit{Due to the nature of this work, some examples contain offensiveness and hate speech. This does not reflect authors’ values, however our aim is to help in detecting and preventing spread of such harmful content.}}\\

The advent of online social networks have created a platform for billions of people to express their thoughts freely on the internet. This has enormous benefits for advancing culture. However, it also can be used by malicious actors to distribute misinformation and offensive content. This led to an increasing interest in the NLP community for the automatic detection of \textit{Hate Speech} (HS) \cite{waseem-hovy-2016-hateful,schmidt-wiegand-2017-survey,zampieri-etal-2019-predicting,macavaney2019hate,league2020online,vogels2021state}. Its dangers are becoming more apparent with studies showing its connection to hate crimes around the globe \cite{maria_hs}. Further, the spread of hateful content on the internet has also been linked to degenerate effects on peoples' psychological well-being \cite{Glati2010TheEO,harmhs}. 

HS is defined as any kind of abusive or offensive language (e.g. insults, threats, etc.) that expresses prejudice against a specific person or a group based on common characteristics such as race, religion or sexual orientation \cite{Davidson2017AutomatedHS,Mollas2020ETHOSAO}. Despite the growing body of HS research, few have focused on it in the context of the Arabic language.

Arabic is the mother tongue of more than 420M people, and is spoken in the fastest growing markets \cite{tinsley_board}. 
Arabic content is rapidly growing on the internet during the past couple of years \cite{abdelali-etal-2021-qadi}. For instance, studies have shown that there are more than 27 million tweets per day in the Arab region \cite{Alshehri2018ThinkBY}. 

\begin{figure}
    \centering
    \includegraphics[width=0.7\linewidth]{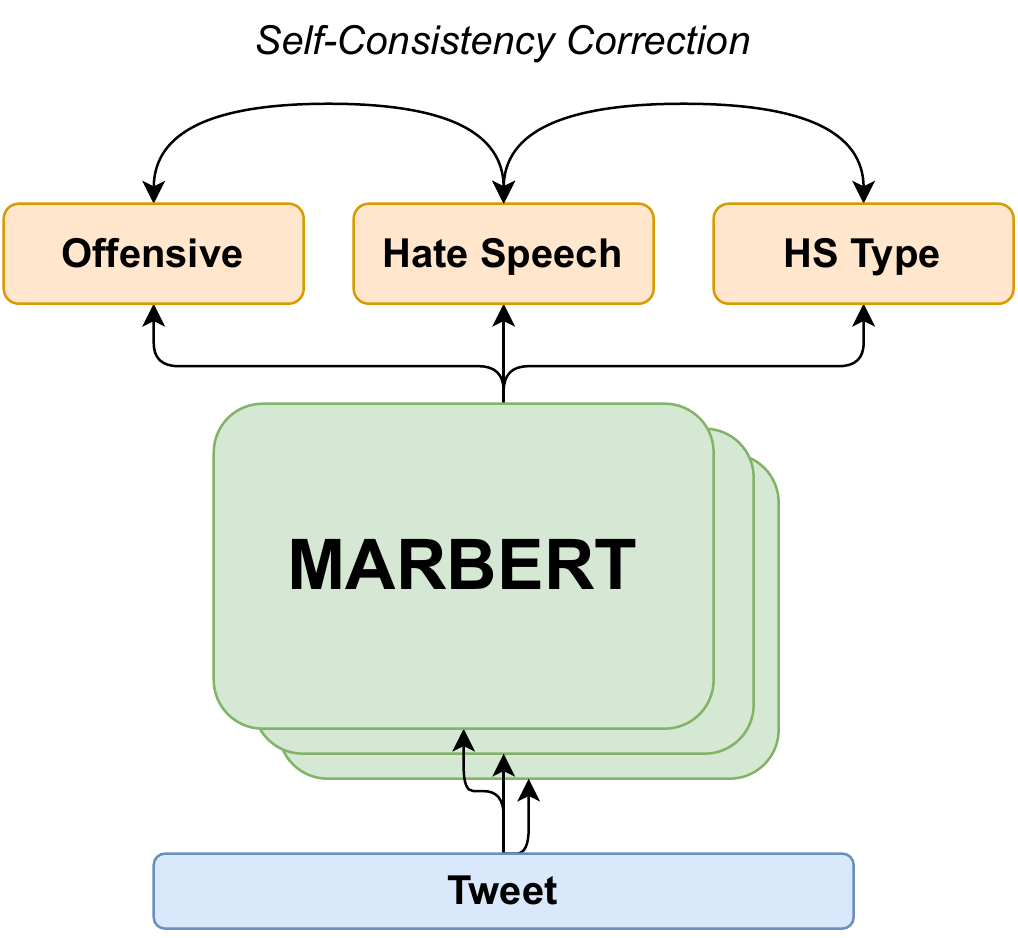}
    \caption{\textbf{System Architecture} The input tweet is encoded using a fine-tuned MARBERT model and the output embedding is given to $3$ task-specific classifiers. The final prediction is computed using an ensemble of those models.}
    \label{fig:model}
\end{figure}

In this work, we focus on the Arabic language by participating in the three subtasks of the \textit{Arabic Fine-Grained Hate Speech Detection} shared task \cite{mubarak2022emojis}. The three subtasks use the same dataset from \cite{mubarak2022emojis} (see Section \ref{sec:dataset} for more details.) The goal of the first subtask is to detect whether a tweet is offensive or not, while the second subtask focuses on HS detection. The third subtask further classifies a HS post into one of six fine-grained categories: Race/Ethnicity/Nationality, Religion/Belief, Ideology, Disability/Disease, Social Class, and Gender. Table \ref{tab:examples} shows an example with its corresponding label for each subtask. An offensive post is not necessarily HS, while a HS post is always offensive. If offensive speech is not targeting an individual or a group based on common characteristics, then it is not HS. 

The contributions of this paper are as follows: (1) We present a solution that outperforms the baseline models of \cite{mubarak2022emojis} on every reported metric; (2) We propose the self-consistency correction method that improves the fine-grained HS subtask even further; (3) We conduct an ablation study and further analysis illustrating the importance of multi-task learning for Arabic HS detection.

\begin{table*}[ht]
\centering
\begin{tabular}{@{}lc@{}}
\toprule
\textbf{Class}  & \textbf{Example}                                                                                                                                             \\ \midrule
\textbf{Clean}             & 
\begin{tabular}[c]{@{}c@{}}
1. \RL{لن تحصل على غدٍ افضل ما دمت تفكر بالامس} (Non-Off)\\
You won’t have a better tomorrow as long you are thinking about yesterday.
\end{tabular} \\ \midrule
\textbf{Offensive}         &   
\begin{tabular}[c]{@{}c@{}}
2. \RL{يلعن ابوك على هالسؤال. عساه ينقرض الكريه. } (Off/Non-HS)\\
May your father be damned for this question! I hope this fool will just wither!
\end{tabular} \\ \midrule
\textbf{Hate Speech}       &                                                 \begin{tabular}[c]{@{}c@{}}
3. \RL{لذا القزم طلعت جايزتن له بس ماعرف يعبر } (Off/HS)\\ This dwarf got two prizes, but he does not know how to express.
\end{tabular} 
\\ \bottomrule
\end{tabular}
\caption{Here we show examples and their translation in English adapted from \protect\cite{mubarak2022emojis}. Example 1 is non offensive (Non-Off), Example 2 is Offensive but not Hate speech (Off/Non-HS), Example 3 is offensive and hate speech (Off/HS). }
\label{tab:examples}
\end{table*}

\section{Motivation}

There has been a growing body of research in recent years for the automatic detection of offensive language and HS online \cite{waseem-hovy-2016-hateful,Davidson2017AutomatedHS,schmidt-wiegand-2017-survey,fortuna2018survey,founta2018large}. Studies have shown that 41\% of internet users have been harassed online with a third of these cases being targeted for something related to their inherent identity such as race or sexual orientation \cite{vogels2021state,league2020online}. The massive amount of content shared on social media platforms renders manual filtering out of such malicious content impossible, driving platform providers to resort to automated means for detecting hateful content. 
On the other hand, machine learning based methods are data hungry and require large amounts of labelled data in order to train  reliable HS classification systems. Moreover, such data has been proven hard to collect especially for low-resource languages such as Arabic. For example, \cite{mubarak-etal-2017-abusive} show that only 1-2\% of a randomly collected sample of Arabic tweets are abusive, and only a small percentage of these are considered HS. Therefore, generalizable and robust systems for detecting offensive and HS content are direly needed. 

Previous work has framed this problem as a binary classification task. However, binary judgments of HS are known to be unreliable \cite{sanguinetti-etal-2018-italian,assimakopoulos2020annotating}. Therefore, in order to collect higher quality HS datasets researchers resorted to more complex annotation schema. For example, \cite{sap2020socialbiasframes,assimakopoulos-etal-2020-annotating} proposed to decompose a post into several subtasks (such as the HS class and group  targeted) in an effort to minimize subjectivity when deciding the HS label.

Here, we leverage the task decomposed dataset provided by \cite{mubarak2022emojis} to train an Arabic transformer in a multitask manner for improving the performance of fine-grained HS detection. 

\section{Dataset}
\label{sec:dataset}

We use the dataset from \cite{mubarak2022emojis}. It consists of $\sim13$k tweets in both Modern Standard Arabic (MSA) and Dialectal forms of Arabic (DA). It is the largest annotated corpus of Arabic tweets that is not biased towards specific topics, genres, or dialects \cite{mubarak2022emojis}. Each tweet was judged by $3$ annotators using crowd-sourcing. Table \ref{tab:stats} shows the number and percentages of each annotated category. The data was split into $70$\% for training, $10$\% for development, and $20$\% for test. The dataset has also annotations for vulgar and violent tweets representing $1.5$\% and $0.7$\% of the whole corpus, respectively, however we are not using them in this work. Moreover, one or more user mentions are reduced to \code{@USER}, URLs are replaced with \code{URL}, and empty lines in original tweets are replaced with \code{<LF>}. See Table \ref{tab:examples} for an example of each annotated class.

One limitation of this dataset is that the classes are highly imbalanced. Moreover, the \textit{Disability/Disease} subclass does not exist in the training set. There are only 3 tweets related to this category and they appear in the validation and test sets only.

\begin{table}[ht]
\centering
\begin{tabular}{@{}lrr@{}}
\toprule
\textbf{Class - Subclass}  & \textbf{\# of Tweets} & \textbf{Percentage (\%)} \\ \midrule
\textbf{Clean}             & 8,235                 & 64.85\%                  \\
\textbf{Offensive}         & 4,463                 & 35.15\%                  \\
\hline
\textbf{Hate Speech}       & 1,339                 & 10.54\%                  \\ 
\textbf{HS - Gender}       & 641                   & 5.05\%                   \\
\textbf{HS - Race}         & 366                   & 2.88\%                   \\
\textbf{HS - Ideology}     & 190                   & 1.50\%                   \\
\textbf{HS - Social Class} & 101                   & 0.80\%                   \\
\textbf{HS - Religion}     & 38                    & 0.30\%                   \\
\textbf{HS - Disability}   & 3                     & 0.02\%                   \\ \bottomrule
\end{tabular}
\caption{\textbf{Dataset Statistics} The number and percentages of tweets as represented in the entire corpus of each annotated category used in this work as described in \protect\cite{mubarak2022emojis}.\protect\footnotemark}
\label{tab:stats}
\end{table}

\footnotetext{The percentages do not add up to 100 since all HS tweets are a subset of the offensive ones.}

\begin{table*}[ht]
\centering
\begin{tabular}{@{}lccccc@{}}
\toprule
\textbf{Subtask}                      & \textbf{Model} & \textbf{Accuracy} & \textbf{Precision} & \textbf{Recall} & \textbf{F1 Macro} \\ \midrule
\multirow{2}{*}{\textbf{OFFD}}   & QARiB       & 84.0\%           & 82.5\%            & 82.1\%         & 82.3\%           \\
                                      & AraHS           & \textbf{86.0\%}              & \textbf{84.6\%}             & \textbf{84.3\%}          & \textbf{84.5\%}            \\
\multirow{2}{*}{\textbf{HSD}} & QARiB       & 93.0\%           & 83.0\%            & 77.7\%         & 80.0\%           \\
                                      & AraHS           & \textbf{94.1\%}           & \textbf{87.0\%}               & \textbf{79.5\%}         & \textbf{82.7\%}            \\
\textbf{HSC}                               & AraHS           & 92.6\%            & 55.1\%             & 50.8\%          & 51.9\%            \\ \bottomrule
\end{tabular}
\caption{Performance on Test for each of the subtasks Offensive Detection (OFFD), Hate Speech Detection (HSD), Hate Speech Classification (HSC) in comparison with the QARiB baseline models reported in \protect\cite{mubarak2022emojis}. Our model, AraHS, outperforms the baseline QARiB on every metric. \textbf{NB:} no baseline was reported for the HSC subtask.}
\label{tab:results}
\end{table*}

\section{System Description}

In this work, we use MARBERTv2 \cite{abdul-mageed-etal-2021-arbert} as our core model. It is publicly available in the HuggingFace library \cite{Wolf2019HuggingFace}. It is pretrained with $1$B multi-dialectal Arabic (DA) tweets which includes both MSA and DA. MARBERTv2 has the same architecture as BERT\textsubscript{BASE} \cite{devlin_etal_2019_bert} with $\sim 163$M parameters, similarly using WordPiece tokenization \cite{wu_2016_wordpiece}. 

We frame the 3 subtasks as a multi-task classification problem. Specifically, the input text is encoded using MARBERTv2 and is then passed to $3$ task-specific classification heads as shown in Figure \ref{fig:model}. Each class specific head is made up of a multi-layered feed forward neural network with layer normalization \cite{Ba2016LayerN}. Concretely, the \texttt{[CLS]} embedding of the final MARBERTv2 transformer block is forwarded to a dense layer with $768$ units, which is then passed through a GELU activation function \cite{Hendrycks2016BridgingNA}, the output of which is normalized using layer normalization, and this is finally given to a linear layer that maps it to the corresponding number of classes. 

The final model is an ensemble of several trained models each of which uses a different set of hyperparameters. 
To obtain the final prediction we perform element-wise multiplication of the corresponding probabilities across the different models then take the argmax.

\paragraph{Self-Consistency Correction} Since we are training one model for all three subtasks, and the subtasks themselves are interdependent, we  leverage that to our advantage. We perform a post-processing step where errors of one classification head are corrected by the others. Concretely, the fine-grained HS prediction is corrected in the following cases: if the tweet is predicted to be offensive and contains HS using the first two classification heads respectively, while the fine-grained classifier predicted that it is not HS. In that case, we take the second most probable class prediction as the label since there is an inconsistency. The other scenario in which it is corrected is when the tweet is predicted as not offensive and does not contain HS while the fine-grained classifier predicted it as one of the HS classes. 


\section{Experimental Setup}

To train the AraHS model, we use the AdamW optimzier \cite{adamw} and a learning rate scheduler that is warmed-up linearly for $500$ steps to some initial learning rate. This is then decayed linearly to zero over the course of $10$ epochs. The model is evaluated on the validation-set $4$ times every epoch with equal intervals, and a checkpoint for the corresponding subtask is saved when its F1-macro score improves. The objective function is the sum of the negative log-likelihood of the three classification heads. The tokenizer encodes the input text using a maximum length of $256$ tokens. The model is trained $12$ times over a grid of $\{2, 4, 8, 16\}$ batch-sizes and $\{1e-5, 5e-6, 1e-6\}$ initial learning rates.

For the fine-grained HS detection subtask, we further finetune the best single model on only this subtask, using the same experimental setup described above.

\section{Results}

Table \ref{tab:results} shows the performance on the test-set for each subtask. Our method (AraHS) outperforms the baseline models reported in \cite{mubarak2022emojis} on every metric: Offensive Detection subtask (OFFD) (accuracy: AraHS 86\% vs. QARiB 84\%; F1-Macro: AraHS 84.5\% vs. QARiB 82.3\%); Hate Speech Detection (HSD) subtask (accuracy: AraHS 94.1\% vs. QARiB 93\%; F1-Macro: AraHS 82.7\% vs. QARiB 80\%). Only the HS Classification (HSC) uses the self-consistency correction method. Since the dataset used in this work does not contain the disability class in the training set, the final HSC F1-macro score degrades considerably.

\subsection{Ablation Study}

In order to demonstrate the importance of training the subtasks jointly, we train each subtask on its own. Specifically, Table \ref{tab:ablation} compares the validation performance of each subtask with its multitask counterpart. Performance improves when using multitask learning (MTL) for the HS subtasks. However, for the offensive subtask we observe similar performance to the single-task trained models. Similar to Table \ref{tab:results}, only the HSC is using self-consistency correction, improving the F1-macro score from $54.8\%$ to $56.6\%$.

\begin{table*}[]
\centering
\begin{tabular}{@{}lccccc@{}}
\toprule
\textbf{Subtask}                      & \textbf{Model} & \textbf{Accuracy} & \textbf{Precision} & \textbf{Recall} & \textbf{F1 Macro} \\ \midrule
\multirow{2}{*}{\textbf{OFFD}}   & Single-task    & \textbf{88.7\%}   & \textbf{87.4\%}    & \textbf{86.1\%} & \textbf{86.7\%}   \\
                                      & Multitask      & 88.5\%            & 87.1\%             & \textbf{86.1\%} & 86.6\%            \\
\multirow{2}{*}{\textbf{HSD}} & Single-task    & 95.8\%            & 87.2\%             & 85.7\%          & 86.4\%            \\
                                      & Multitask      & \textbf{96.2\%}   & \textbf{87.7\%}    & \textbf{88.4\%} & \textbf{88.1\%}   \\
\multirow{2}{*}{\textbf{HSC}}     & Single-task    & \textbf{95.3\%}   & \textbf{72.4\%}    & 46.8\%          & 51.0\%            \\
                                      & Multitask      & 95.0/94.4\%            & 58.5/54.8\%             & 52.5/\textbf{58.8\%} & 54.8/\textbf{56.6\%}   \\ \bottomrule
\end{tabular}
\caption{The validation performance on each subtask. The single-task models are trained on the subtask alone, while the multitask model trains all subtasks jointly. The results before and after applying the self-consistency correction as a post-processing step is shown for the HSC subtask. \textbf{Bold} shows the best result for each subtask.}
\label{tab:ablation}
\end{table*}

\section{Error Analysis}

The subtasks we are training are not independent of one another, even though we are modelling them that way. For example, as previously mentioned, a tweet that is considered HS is automatically offensive as well, and needless to say that each of the fine-grained HS classes (e.g. race, gender, etc.) are HS. Therefore, in this section we want to measure the degree of self-consistency within the trained multitask model. Specifically, we take the predictions of the best ensemble of models and calculate the percentage of contradiction between each classification head (see Table \ref{tab:analysis}). Concretely, we compute the number of times in which the OFFD head yielded a negative prediction whereas the HSD or the HSC yielded a positive one. This is a contradiction since each HS post must be offensive as well. Similarly, we calculate the number of times the HSD head predicted negative while the HSC predicted positive and vice versa. As illustrated in Table~\ref{tab:analysis}, correcting the HSC head based on the two other subtasks reduces the contradiction considerably (from 2.6\% to 0.79\%) while achieving a better performance overall.

\begin{table}[ht]
\centering
\begin{tabular}{@{}lc@{}}
\toprule
\textbf{Subtask}     & \textbf{Contradiction (\%)} \\ \midrule
\textbf{OFFD}   & 2.44\%                     \\
\textbf{HSD} & 2.60\%                     \\
\textbf{HSC}     & 2.60\% / 0.79\%                     \\ \bottomrule
\end{tabular}
\caption{Percentage of contradiction between classification heads before and after self-correction for the HSC. Each row correspond to the best checkpoint achieved for that subtask.}
\label{tab:analysis}
\end{table}

Furthermore, Table \ref{tab:dis_examples} shows two examples where the classification heads disagreed with one another. For example, the first tweet was detected as HS but the fine-grained classification head classified it as non-HS leading to a disagreement. Using our self-consistency correction method, the model was able to correct itself and yield the correct label, which was the \textit{Ideology} subclass in this case. Example 1 in the table is a modification of an Arabic adage: ``\texttt{\RL{الطيور على اشكالها تقع}}'', corresponding to ``\texttt{Birds of a feather flock together}. Changing \textit{birds} to \textit{frogs} implies ugliness. Such tweets are not straight forward to classify since they require an understanding of cultural knowledge and implicit social nuance that is not explicitly encoded  in language models such as MARBERT. One way to mitigate this is to finetune the model on a corpus that contains such information explicitly incorporating such inductive bias. Another method would be training the language model to generate the implication of the tweet as an additional subtask. The other example in the table implies that people of a certain nationality are ignorant. We believe that the provided gold label is incorrect (not HS). We believe that this tweet constitutes HS because it is offensive (a certain group of people is ignorant since they parrot rather than understand information) and it targets a group. Accordingly, the model was able to successfully predict it as HS, and even yield the correct class for it using the self-consistency method.

\begin{table*}[]
\centering
\begin{tabular}{@{}cccc@{}}
\toprule
\textbf{Tweet}                           & \textbf{OFFD}      & \textbf{HSD}    & \textbf{HSC}  \\ \midrule
\begin{tabular}[c]{@{}c@{}} \RL{الضفادع على اشكالها تقع}  \\ Frogs settle with their own kind. \end{tabular} & \cmark | \cmark & \cmark | \cmark & \xmark | Ideology \\
\begin{tabular}[c]{@{}c@{}} \RL{أستغفر الله العظيم شعب حافظ مش فاهم} \\ The people of this country memorize without understanding. \end{tabular} & \cmark | \xmark & \cmark | \xmark & \xmark | \xmark   \\ 
\begin{tabular}[c]{@{}c@{}} \RL{ضروري من الطقوس في هذا المجتمع القائم على النفاق الاجتماعي} \\ Necessary rituals in this community that is based on social hypocrisy. \end{tabular} & \xmark | \cmark & \xmark | \cmark & \xmark | Nationality   \\ \bottomrule
\end{tabular}
\caption{The first two are examples that led to a disagreement between the classification heads, while the third one show a false negative example. Below each example is a rough English translation that is used to convey the meaning. On the right-side of the table, the prediction of the model is shown first followed by the ground truth label for each subtask. Note that the self-consistency correction method was able to correct the first two examples among others.}
\label{tab:dis_examples}
\end{table*}

In Table \ref{tab:error_type} we report the percentage of false positives (FP) and false negatives (FN) of the best checkpoint of each subtask. To compute the percentage of FP and FN for the HSC subtask we convert it into a binary variable with negative implying that the prediction is not-HS and positive otherwise.  Interestingly, the self-consistency correction method increases the percentage of FPs, as it takes the second top prediction as its label when both the HSD and OFFD are positive. We note that HS systems can tolerate more false positives (i.e. over enforcement) than false negatives (i.e. under enforcement), since the latter will lead to more propagation of harm. This highlights the advantage of self-consistency correction. 

\begin{table}[]
\centering
\begin{tabular}{@{}lcc@{}}
\toprule
\textbf{}            & \textbf{False +ve (\%)} & \textbf{False -ve (\%)} \\ \midrule
\textbf{OFFD}   & 4.96\%      & 6.54\%          \\
\textbf{HSD} & 1.97\%      & 1.81\%        \\
\textbf{HSC}     & 2.52\%/3.86\% & 2.44\%/1.73\% \\ \bottomrule
\end{tabular}
\caption{Percentage of false positives and false negatives for the best checkpoint for each subtask. In the HSC we report the percentage before and after applying the self-consistency correction method.}
\label{tab:error_type}
\end{table}

\section{Related Work}

\paragraph{Datasets}
The first Arabic HS dataset was collected by \cite{Albadi2018AreTO} and consisted of $\sim 6.6$k Arabic HS tweets. In an effort to collect a more dialect specific dataset, \cite{Haddad19} compiled $6$k tweets of the Tunisian dialect containing both abusive language and HS. \cite{mulki-etal-2019-l} similarly collected a Levantine HS dataset. In the multilingual front, \cite{ousidhoum-etal-2019-multilingual} created a HS dataset made up of $13$k Arabic, English and French tweets with fine-grained labels covering different aspects such as target groups, directness, target attributes and hostility types. 

\paragraph{Models}
Early work tackled this problem by extracting n-gram features using term frequency weighting, which was then passed to a Support Vector Machine (SVM) and Naive Bayes (NB) classifiers \cite{mulki-etal-2019-l}. Other work used a gated recurrent unit (GRU) coupled with an SVM trained on the AraVec embeddings \cite{Ashi2018PretrainedWE} to classify HS \cite{Albadi2018AreTO}. \cite{hassan-etal-2020-alt} used an ensemble of SVM, CNN-BiLSTM and feed-forward neural networks for HS detection. \cite{Duwairi2021ADL} showed that CNN models outperform their CNN-LSTM and CNN-BiLSTM counterparts in detecting HS when treated as a binary classification task on the ArHS dataset. 

\paragraph{Multitask Learning}
In the Offensive Detection shared task co-located with the 4th Workshop on Open-Source Arabic Corpora and Processing Tools (OSACT4) \cite{osact-2020-open}, \cite{djandji-etal-2020-multi} trained AraBERT \cite{antoun-etal-2020-arabert} on multiple tasks simultaneously achieving the best score on the shared task. \cite{el-mahdaouy-etal-2021-deep} used a model based on MARBERT that employed MTL. In another line of work, a CNN-BiLSTM based architecture was trained using MTL to detect HS and OFF language \cite{abu-farha-magdy-2020-multitask}. That model used extra sentiment information \cite{abu-farha-magdy-2019-mazajak} during training. Finally, \cite{Aldjanabi21} explore a multi-corpus-based learning approach built on top of MARBERT. It uses MTL from three datasets for improving OFF and HS detection. Unlike previous work, our paper focuses on improving fine-grained Arabic HS classification using MTL and self-consistency correction on the new dataset introduced in \cite{mubarak2022emojis}. 

\section{Conclusion}

In this paper, we propose MTL as an approach to Hate Speech Classification. Our proposed model, AraHS, outperforms the baseline models. AraHS is an ensemble of MARBERT \cite{abdul-mageed-etal-2021-arbert} models trained with different hyperparameters using MTL. The fine-grained HS subtask is then finetuned on its own for a couple of epochs. We demonstrate the importance of training the three subtasks jointly through an ablation study and propose the self-consistency correction method that improves the final result even further. In future work, we would like to explore the limits of combining multilingual models (e.g. mBART \cite{Liu2020MultilingualDP}) with Arabic monolingual models such as MARBERT. Further, we would like to explore treating the problem as a conditional generation task using the AraT5 model \cite{Nagoudi2021AraT5TT} that has been shown to outperform MARBERT on the Arabic language understanding evaluation benchmark (ARLUE) \cite{abdul-mageed-etal-2021-arbert}.

\section{Ethics and Social Impact}

Modern deep learning models are energy intensive and can cause environmental damage due to the carbon dioxide emissions required for running modern hardware. Studies have shown that training a BERT model on GPU has a comparable carbon footprint to a trans-American flight \cite{Strubell2019EnergyAP}. In this work, even though we do not pre-train the model, we still run multiple experiments across a grid of hyperparameters, that when combined consumes significant energy. Therefore, one of the reasons we chose multitask learning (MTL) is that we can reduce the amount of training substantially by training one model on multiple tasks. MTL does not only offer energy efficiency, but is also more data efficient, it has been shown to converge faster by leveraging auxiliary information and reduces over-fitting through shared representations \cite{Crawshaw2020MultiTaskLW}.

Further, building models for detecting OFF language and HS can help improve the moderation of hateful content on the internet. This can potentially lead to less hate crimes and better psychological well-being for users receiving such content. However, the authors are aware of potential misuse of HS models, such as propagating the spread of HS rather than suppressing it. Therefore, human moderation is required for preventing such misuse.


\section{Bibliographical References}
\label{reference}

\bibliographystyle{lrec2022-bib}
\bibliography{references}

\begin{thebibliography}{}

\bibitem[\protect\citename{Abdelali \bgroup et al.\egroup
  }2021]{abdelali-etal-2021-qadi}
Abdelali, A., Mubarak, H., Samih, Y., Hassan, S., and Darwish, K.
\newblock (2021).
\newblock {QADI}: {A}rabic dialect identification in the wild.
\newblock In {\em Proceedings of the Sixth Arabic Natural Language Processing
  Workshop}, pages 1--10, Kyiv, Ukraine (Virtual), April. Association for
  Computational Linguistics.

\bibitem[\protect\citename{Abdul-Mageed \bgroup et al.\egroup
  }2021]{abdul-mageed-etal-2021-arbert}
Abdul-Mageed, M., Elmadany, A., and Nagoudi, E. M.~B.
\newblock (2021).
\newblock {ARBERT} {\&} {MARBERT}: Deep bidirectional transformers for
  {A}rabic.
\newblock In {\em Proceedings of the 59th Annual Meeting of the Association for
  Computational Linguistics and the 11th International Joint Conference on
  Natural Language Processing (Volume 1: Long Papers)}, pages 7088--7105,
  Online, August. Association for Computational Linguistics.

\bibitem[\protect\citename{Abu~Farha and
  Magdy}2019]{abu-farha-magdy-2019-mazajak}
Abu~Farha, I. and Magdy, W.
\newblock (2019).
\newblock {M}azajak: An online {A}rabic sentiment analyser.
\newblock In {\em Proceedings of the Fourth Arabic Natural Language Processing
  Workshop}, pages 192--198, Florence, Italy, August. Association for
  Computational Linguistics.

\bibitem[\protect\citename{Abu~Farha and
  Magdy}2020]{abu-farha-magdy-2020-multitask}
Abu~Farha, I. and Magdy, W.
\newblock (2020).
\newblock Multitask learning for {A}rabic offensive language and hate-speech
  detection.
\newblock In {\em Proceedings of the 4th Workshop on Open-Source Arabic Corpora
  and Processing Tools, with a Shared Task on Offensive Language Detection},
  pages 86--90, Marseille, France, May. European Language Resource Association.

\bibitem[\protect\citename{Al-Khalifa \bgroup et al.\egroup
  }2020]{osact-2020-open}
Al-Khalifa, H., Magdy, W., Darwish, K., Elsayed, T., and Mubarak, H.
\newblock (2020).
\newblock Proceedings of the 4th workshop on open-source arabic corpora and
  processing tools, with a shared task on offensive language detection.
\newblock Marseille, France, May. European Language Resource Association.

\bibitem[\protect\citename{Albadi \bgroup et al.\egroup }2018]{Albadi2018AreTO}
Albadi, N., Kurdi, M., and Mishra, S.
\newblock (2018).
\newblock Are they our brothers? analysis and detection of religious hate
  speech in the arabic twittersphere.
\newblock {\em 2018 IEEE/ACM International Conference on Advances in Social
  Networks Analysis and Mining (ASONAM)}, pages 69--76.

\bibitem[\protect\citename{Aldjanabi \bgroup et al.\egroup }2021]{Aldjanabi21}
Aldjanabi, W., Dahou, A., Al-qaness, M. A.~A., Elsayed Abd~Elaziz, M., Helmi,
  A., and Damasevicius, R.
\newblock (2021).
\newblock Arabic offensive and hate speech detection using a cross-corpora
  multi-task learning model.
\newblock {\em Informatics}, 8:69, 10.

\bibitem[\protect\citename{Alshehri \bgroup et al.\egroup
  }2018]{Alshehri2018ThinkBY}
Alshehri, A., Nagoudi, E. M.~B., Alhuzali, H., and Abdul-Mageed, M.
\newblock (2018).
\newblock Think before your click: Data and models for adult content in arabic
  twitter.

\bibitem[\protect\citename{Antoun \bgroup et al.\egroup
  }2020]{antoun-etal-2020-arabert}
Antoun, W., Baly, F., and Hajj, H.
\newblock (2020).
\newblock {A}ra{BERT}: Transformer-based model for {A}rabic language
  understanding.
\newblock In {\em Proceedings of the 4th Workshop on Open-Source Arabic Corpora
  and Processing Tools, with a Shared Task on Offensive Language Detection},
  pages 9--15, Marseille, France, May. European Language Resource Association.

\bibitem[\protect\citename{Ashi \bgroup et al.\egroup
  }2018]{Ashi2018PretrainedWE}
Ashi, M.~M., Siddiqui, M.~A., and Nadeem, F.
\newblock (2018).
\newblock Pre-trained word embeddings for arabic aspect-based sentiment
  analysis of airline tweets.
\newblock In {\em AISI}.

\bibitem[\protect\citename{Assimakopoulos \bgroup et al.\egroup
  }2020a]{assimakopoulos2020annotating}
Assimakopoulos, S., Muskat, R.~V., van~der Plas, L., and Gatt, A.
\newblock (2020a).
\newblock Annotating for hate speech: The maneco corpus and some input from
  critical discourse analysis.
\newblock {\em arXiv preprint arXiv:2008.06222}.

\bibitem[\protect\citename{Assimakopoulos \bgroup et al.\egroup
  }2020b]{assimakopoulos-etal-2020-annotating}
Assimakopoulos, S., Vella~Muskat, R., van~der Plas, L., and Gatt, A.
\newblock (2020b).
\newblock Annotating for hate speech: The {M}a{N}e{C}o corpus and some input
  from critical discourse analysis.
\newblock In {\em Proceedings of the 12th Language Resources and Evaluation
  Conference}, pages 5088--5097, Marseille, France, May. European Language
  Resources Association.

\bibitem[\protect\citename{Ba \bgroup et al.\egroup }2016]{Ba2016LayerN}
Ba, J., Kiros, J., and Hinton, G.~E.
\newblock (2016).
\newblock Layer normalization.
\newblock {\em ArXiv}, abs/1607.06450.

\bibitem[\protect\citename{Crawshaw}2020]{Crawshaw2020MultiTaskLW}
Crawshaw, M.
\newblock (2020).
\newblock Multi-task learning with deep neural networks: A survey.
\newblock {\em ArXiv}, abs/2009.09796.

\bibitem[\protect\citename{Davidson \bgroup et al.\egroup
  }2017]{Davidson2017AutomatedHS}
Davidson, T., Warmsley, D., Macy, M.~W., and Weber, I.
\newblock (2017).
\newblock Automated hate speech detection and the problem of offensive
  language.
\newblock In {\em ICWSM}.

\bibitem[\protect\citename{Devlin \bgroup et al.\egroup
  }2019]{devlin_etal_2019_bert}
Devlin, J., Chang, M.-W., Lee, K., and Toutanova, K.
\newblock (2019).
\newblock {BERT}: Pre-training of deep bidirectional transformers for language
  understanding.
\newblock In {\em Proceedings of the 2019 Conference of the North {A}merican
  Chapter of the Association for Computational Linguistics: Human Language
  Technologies, Volume 1 (Long and Short Papers)}, pages 4171--4186,
  Minneapolis, Minnesota, June. Association for Computational Linguistics.

\bibitem[\protect\citename{Djandji \bgroup et al.\egroup
  }2020]{djandji-etal-2020-multi}
Djandji, M., Baly, F., Antoun, W., and Hajj, H.
\newblock (2020).
\newblock Multi-task learning using {A}ra{B}ert for offensive language
  detection.
\newblock In {\em Proceedings of the 4th Workshop on Open-Source Arabic Corpora
  and Processing Tools, with a Shared Task on Offensive Language Detection},
  pages 97--101, Marseille, France, May. European Language Resource
  Association.

\bibitem[\protect\citename{Duwairi \bgroup et al.\egroup }2021]{Duwairi2021ADL}
Duwairi, R., Hayajneh, A., and Quwaider, M.
\newblock (2021).
\newblock A deep learning framework for automatic detection of hate speech
  embedded in arabic tweets.
\newblock {\em Arabian Journal for Science and Engineering}, 46:1--14.

\bibitem[\protect\citename{El~Mahdaouy \bgroup et al.\egroup
  }2021]{el-mahdaouy-etal-2021-deep}
El~Mahdaouy, A., El~Mekki, A., Essefar, K., El~Mamoun, N., Berrada, I., and
  Khoumsi, A.
\newblock (2021).
\newblock Deep multi-task model for sarcasm detection and sentiment analysis in
  {A}rabic language.
\newblock In {\em Proceedings of the Sixth Arabic Natural Language Processing
  Workshop}, pages 334--339, Kyiv, Ukraine (Virtual), April. Association for
  Computational Linguistics.

\bibitem[\protect\citename{Fortuna and Nunes}2018]{fortuna2018survey}
Fortuna, P. and Nunes, S.
\newblock (2018).
\newblock A survey on automatic detection of hate speech in text.
\newblock {\em ACM Computing Surveys (CSUR)}, 51(4):1--30.

\bibitem[\protect\citename{Founta \bgroup et al.\egroup }2018]{founta2018large}
Founta, A.~M., Djouvas, C., Chatzakou, D., Leontiadis, I., Blackburn, J.,
  Stringhini, G., Vakali, A., Sirivianos, M., and Kourtellis, N.
\newblock (2018).
\newblock Large scale crowdsourcing and characterization of twitter abusive
  behavior.
\newblock In {\em Twelfth International AAAI Conference on Web and Social
  Media}.

\bibitem[\protect\citename{G{\"u}laçti}2010]{Glati2010TheEO}
G{\"u}laçti, F.
\newblock (2010).
\newblock The effect of perceived social support on subjective well-being.
\newblock {\em Procedia - Social and Behavioral Sciences}, 2:3844--3849.

\bibitem[\protect\citename{Haddad \bgroup et al.\egroup }2019]{Haddad19}
Haddad, H., Mulki, H., and Oueslati, A., (2019).
\newblock {\em T-HSAB: A Tunisian Hate Speech and Abusive Dataset}, pages
  251--263.
\newblock 10.

\bibitem[\protect\citename{Hassan \bgroup et al.\egroup
  }2020]{hassan-etal-2020-alt}
Hassan, S., Samih, Y., Mubarak, H., Abdelali, A., Rashed, A., and Chowdhury,
  S.~A.
\newblock (2020).
\newblock {ALT} submission for {OSACT} shared task on offensive language
  detection.
\newblock In {\em Proceedings of the 4th Workshop on Open-Source Arabic Corpora
  and Processing Tools, with a Shared Task on Offensive Language Detection},
  pages 61--65, Marseille, France, May. European Language Resource Association.

\bibitem[\protect\citename{Hendrycks and Gimpel}2016]{Hendrycks2016BridgingNA}
Hendrycks, D. and Gimpel, K.
\newblock (2016).
\newblock Bridging nonlinearities and stochastic regularizers with gaussian
  error linear units.
\newblock {\em ArXiv}, abs/1606.08415.

\bibitem[\protect\citename{League}2020]{league2020online}
League, A.-D.
\newblock (2020).
\newblock Online hate and harassment. the american experience 2021.
\newblock {\em Center for Technology and Society. Retrieved from www. adl.
  org/media/14643/download}.

\bibitem[\protect\citename{Liu \bgroup et al.\egroup
  }2020]{Liu2020MultilingualDP}
Liu, Y., Gu, J., Goyal, N., Li, X., Edunov, S., Ghazvininejad, M., Lewis, M.,
  and Zettlemoyer, L.
\newblock (2020).
\newblock Multilingual denoising pre-training for neural machine translation.
\newblock {\em Transactions of the Association for Computational Linguistics},
  8:726--742.

\bibitem[\protect\citename{Loshchilov and Hutter}2019]{adamw}
Loshchilov, I. and Hutter, F.
\newblock (2019).
\newblock Decoupled weight decay regularization.
\newblock In {\em International Conference on Learning Representations}.

\bibitem[\protect\citename{MacAvaney \bgroup et al.\egroup
  }2019]{macavaney2019hate}
MacAvaney, S., Yao, H.-R., Yang, E., Russell, K., Goharian, N., and Frieder, O.
\newblock (2019).
\newblock Hate speech detection: Challenges and solutions.
\newblock {\em PloS one}, 14(8):e0221152.

\bibitem[\protect\citename{Mollas \bgroup et al.\egroup
  }2020]{Mollas2020ETHOSAO}
Mollas, I., Chrysopoulou, Z., Karlos, S., and Tsoumakas, G.
\newblock (2020).
\newblock Ethos: an online hate speech detection dataset.
\newblock {\em ArXiv}, abs/2006.08328.

\bibitem[\protect\citename{Mubarak \bgroup et al.\egroup
  }2017]{mubarak-etal-2017-abusive}
Mubarak, H., Darwish, K., and Magdy, W.
\newblock (2017).
\newblock Abusive language detection on {A}rabic social media.
\newblock In {\em Proceedings of the First Workshop on Abusive Language
  Online}, pages 52--56, Vancouver, BC, Canada, August. Association for
  Computational Linguistics.

\bibitem[\protect\citename{Mubarak \bgroup et al.\egroup
  }2022]{mubarak2022emojis}
Mubarak, H., Hassan, S., and Chowdhury, S.~A.
\newblock (2022).
\newblock Emojis as anchors to detect arabic offensive language and hate
  speech.
\newblock {\em arXiv preprint arXiv:2201.06723}.

\bibitem[\protect\citename{Mulki \bgroup et al.\egroup
  }2019]{mulki-etal-2019-l}
Mulki, H., Haddad, H., Bechikh~Ali, C., and Alshabani, H.
\newblock (2019).
\newblock {L}-{HSAB}: A {L}evantine {T}witter dataset for hate speech and
  abusive language.
\newblock In {\em Proceedings of the Third Workshop on Abusive Language
  Online}, pages 111--118, Florence, Italy, August. Association for
  Computational Linguistics.

\bibitem[\protect\citename{Nagoudi \bgroup et al.\egroup
  }2021]{Nagoudi2021AraT5TT}
Nagoudi, E. M.~B., Elmadany, A., and Abdul-Mageed, M.
\newblock (2021).
\newblock Arat5: Text-to-text transformers for arabic language generation.

\bibitem[\protect\citename{Ousidhoum \bgroup et al.\egroup
  }2019]{ousidhoum-etal-2019-multilingual}
Ousidhoum, N., Lin, Z., Zhang, H., Song, Y., and Yeung, D.-Y.
\newblock (2019).
\newblock Multilingual and multi-aspect hate speech analysis.
\newblock In {\em Proceedings of the 2019 Conference on Empirical Methods in
  Natural Language Processing and the 9th International Joint Conference on
  Natural Language Processing (EMNLP-IJCNLP)}, pages 4675--4684, Hong Kong,
  China, November. Association for Computational Linguistics.

\bibitem[\protect\citename{Paz \bgroup et al.\egroup }2020]{maria_hs}
Paz, M.~A., Montero-Díaz, J., and Moreno-Delgado, A.
\newblock (2020).
\newblock Hate speech: A systematized review.
\newblock {\em SAGE Open}, 10(4):2158244020973022.

\bibitem[\protect\citename{Sanguinetti \bgroup et al.\egroup
  }2018]{sanguinetti-etal-2018-italian}
Sanguinetti, M., Poletto, F., Bosco, C., Patti, V., and Stranisci, M.
\newblock (2018).
\newblock An {I}talian {T}witter corpus of hate speech against immigrants.
\newblock In {\em Proceedings of the Eleventh International Conference on
  Language Resources and Evaluation ({LREC} 2018)}, Miyazaki, Japan, May.
  European Language Resources Association (ELRA).

\bibitem[\protect\citename{Sap \bgroup et al.\egroup
  }2020]{sap2020socialbiasframes}
Sap, M., Gabriel, S., Qin, L., Jurafsky, D., Smith, N.~A., and Choi, Y.
\newblock (2020).
\newblock Social bias frames: Reasoning about social and power implications of
  language.
\newblock In {\em ACL}.

\bibitem[\protect\citename{Schmidt and
  Wiegand}2017]{schmidt-wiegand-2017-survey}
Schmidt, A. and Wiegand, M.
\newblock (2017).
\newblock A survey on hate speech detection using natural language processing.
\newblock In {\em Proceedings of the Fifth International Workshop on Natural
  Language Processing for Social Media}, pages 1--10, Valencia, Spain, April.
  Association for Computational Linguistics.

\bibitem[\protect\citename{Strubell \bgroup et al.\egroup
  }2019]{Strubell2019EnergyAP}
Strubell, E., Ganesh, A., and McCallum, A.
\newblock (2019).
\newblock Energy and policy considerations for deep learning in nlp.
\newblock {\em ArXiv}, abs/1906.02243.

\bibitem[\protect\citename{Tinsley and Board}2013]{tinsley_board}
Tinsley, T. and Board, K.
\newblock (2013).
\newblock {\em {L}anguages for the {F}uture}.
\newblock British Council.

\bibitem[\protect\citename{Vogels}2021]{vogels2021state}
Vogels, E.~A.
\newblock (2021).
\newblock The state of online harassment.
\newblock {\em Pew Research Center}, 13.

\bibitem[\protect\citename{Waldron}2012]{harmhs}
Waldron, J.
\newblock (2012).
\newblock {\em The Harm in Hate Speech}.
\newblock Harvard University Press.

\bibitem[\protect\citename{Waseem and Hovy}2016]{waseem-hovy-2016-hateful}
Waseem, Z. and Hovy, D.
\newblock (2016).
\newblock Hateful symbols or hateful people? predictive features for hate
  speech detection on {T}witter.
\newblock In {\em Proceedings of the {NAACL} Student Research Workshop}, pages
  88--93, San Diego, California, June. Association for Computational
  Linguistics.

\bibitem[\protect\citename{Wolf \bgroup et al.\egroup
  }2019]{Wolf2019HuggingFace}
Wolf, T., Debut, L., Sanh, V., Chaumond, J., Delangue, C., Moi, A., Cistac, P.,
  Rault, T., Louf, R., Funtowicz, M., and Brew, J.
\newblock (2019).
\newblock Huggingface's transformers: State-of-the-art natural language
  processing.
\newblock {\em ArXiv}, abs/1910.03771.

\bibitem[\protect\citename{Wu \bgroup et al.\egroup }2016]{wu_2016_wordpiece}
Wu, Y., Schuster, M., Chen, Z., Le, Q., Norouzi, M., Macherey, W., Krikun, M.,
  Cao, Y., Gao, Q., Macherey, K., Klingner, J., Shah, A., Johnson, M., Liu, X.,
  Kaiser, u., Gouws, S., Kato, Y., Kudo, T., Kazawa, H., and Dean, J.
\newblock (2016).
\newblock Google's neural machine translation system: Bridging the gap between
  human and machine translation.
\newblock 09.

\bibitem[\protect\citename{Zampieri \bgroup et al.\egroup
  }2019]{zampieri-etal-2019-predicting}
Zampieri, M., Malmasi, S., Nakov, P., Rosenthal, S., Farra, N., and Kumar, R.
\newblock (2019).
\newblock Predicting the type and target of offensive posts in social media.
\newblock In {\em Proceedings of the 2019 Conference of the North {A}merican
  Chapter of the Association for Computational Linguistics: Human Language
  Technologies, Volume 1 (Long and Short Papers)}, pages 1415--1420,
  Minneapolis, Minnesota, June. Association for Computational Linguistics.

\end{thebibliography}

\end{document}